\documentclass[11pt]{article}

\usepackage[final]{acl} 

\usepackage{times}
\usepackage{latexsym}
\usepackage[T1]{fontenc}
\usepackage[utf8]{inputenc}
\usepackage{microtype}
\usepackage{inconsolata}
\usepackage{graphicx}
\usepackage{multirow}
\usepackage{booktabs}
\usepackage{tikz}
\usetikzlibrary{shapes.geometric, arrows.meta, positioning, fit, backgrounds}
\usepackage{placeins}
\usepackage{float}

\title{TruthSplit: Operationalizing Conditional Validity in Arguments \\Through Multi-Perspective Reasoning}

\author{
\begin{tabular}{c}
Benjamin Stieger\thanks{Equal contribution.}$^{1}$ \quad
Maximilian Terberger\footnotemark[1]$^{1}$ \quad
Thomas Huber$^{2}$ \quad
Christina Niklaus$^{2}$ \\
University of St.~Gallen \\
$^{1}$\texttt{\{benjamin.stieger,maximilian.terberger\}@student.unisg.ch} \\
$^{2}$\texttt{\{thomas.huber,christina.niklaus\}@unisg.ch}
\end{tabular}
}

\begin{document}

\maketitle

\begin{abstract}
We present TruthSplit, an interactive system for multi-perspective argument analysis. Existing argumentation tools typically analyze properties of the argument itself, such as structure, quality, stance, or persuasiveness, while leaving perspective-specific background knowledge implicit. TruthSplit addresses this gap by supporting an exploratory analysis of how the same claim can lead to different conclusions when interpreted through worldview-specific values, assumptions, and conceptual definitions. We refer to this perspective-dependent analysis as conditional validity.
Given an input argumentative text, TruthSplit extracts claims and premises, applies a three-layer natural language inference (NLI) approach to assess both logical and worldview-specific normative consistency, and conditions large language model (LLM) reasoning on structured worldview profiles that encode core values and decision principles. The system then generates perspective-specific interpretations, identifies value conflicts and assumption gaps, and visualizes divergence through interactive analytical interfaces. 
\end{abstract}

\section{Introduction}

In an era of increasing polarization~\citep{iyengar2019}, understanding disagreement has become crucial. Consider universal basic income (UBI): one person argues ``I oppose UBI because it undermines individual responsibility,'' while another maintains ``I support UBI as it can provide individuals with the financial security to engage in sustainable practices, which is good for the environment.'' Both may have examined the same data yet reach opposing conclusions~\citep{mercier2011}. Traditional argumentation tools focus on identifying the structure \citep{10.1145/1568234.1568246} or quality of arguments \citep{wachsmuth-etal-2017-argumentation,wachsmuth-etal-2024-argument}, but fail to address: \textit{How can the same argument be valid from multiple, seemingly incompatible perspectives?}

This is a difficult problem because disagreement often stems not from flawed reasoning, but from deeper structural differences: different assumptions about how the world works~\citep{tetlock2005,kuhn1962}, distinct value priorities~\citep{berlin1969,haidt2012}, and varying definitions of contested societal concepts like ``freedom'' or ``justice''~\citep{rawls1993}. In the previous example, one might prioritize individual liberty (``freedom'' as freedom \textit{to} move---positive freedom), while others prioritize collective security (``freedom'' as freedom \textit{from} threats---negative freedom)~\citep{berlin1969,snyder2018}. Both positions may be logically consistent within their ideological frameworks, yet incompatible under a single standard.

\begin{figure*}[t]
\centering
\includegraphics[width=1.0\textwidth]{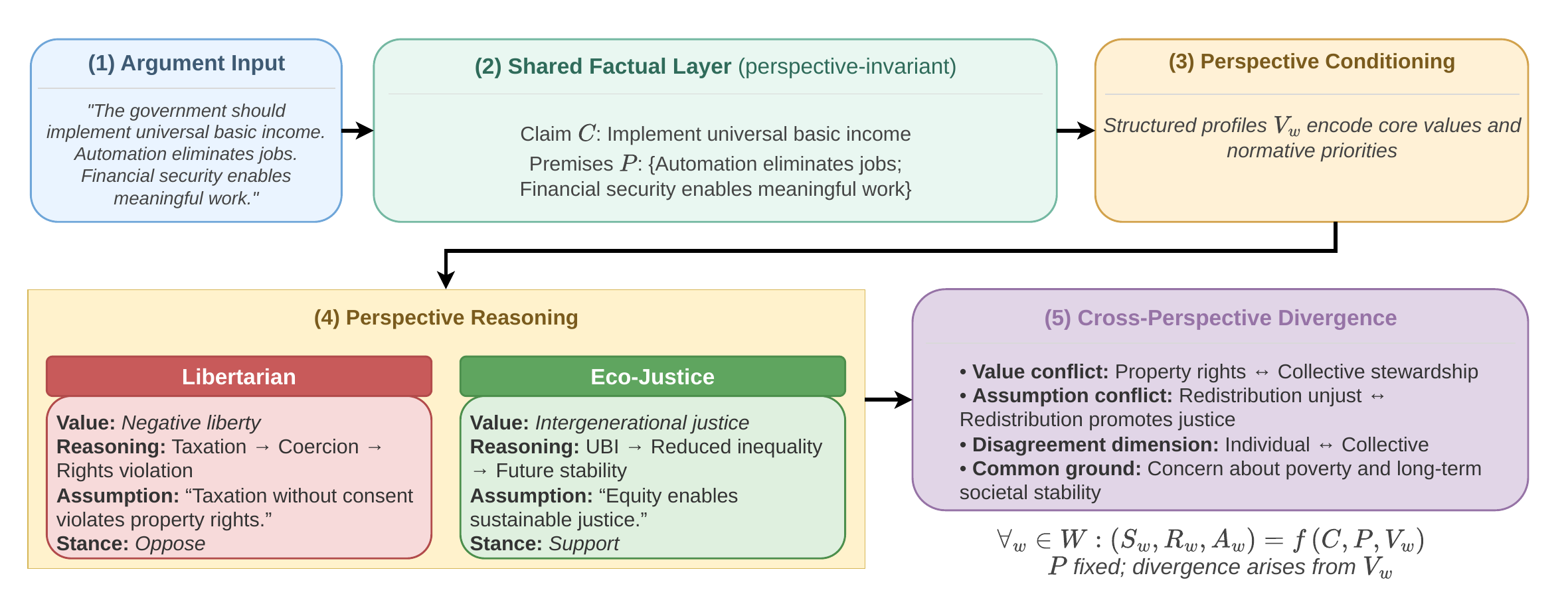}
\caption{Conceptual overview of TruthSplit. An input argument is decomposed into a claim $C$ and premises $P$, forming a shared factual layer that is invariant across perspectives. Structured worldview profiles $V_w$ condition reasoning generation, producing a reasoning layer $R_w$, an assumption layer $A_w$, and a stance $S_w$ for each worldview. Cross-worldview divergence analysis identifies value conflicts, assumption conflicts, disagreement dimensions, and common ground. Disagreement emerges from differences in normative priors rather than factual inconsistency.}
\label{fig:conceptual_overview}
\end{figure*}

This points to \textit{conditional validity}: the idea that the validity of a conclusion depends not only on the premises from which it is drawn, but also on the values, assumptions, and conceptual definitions through which it is interpreted~\citep{rawls1993,lougheed2021}. TruthSplit treats this as an exploratory system concept rather than a formal theory of logical validity, using explicit worldview profiles to compare how different perspectives evaluate the same claim-premise relations.
Existing tools fail because they assume universal correctness. Argumentation systems classify arguments as ``correct'' or ``fallacious''~\citep{goffredo-etal-2023-argument} but cannot capture how an argument might be valid from one perspective yet invalid from another. Prior work has focused on identifying argumentative structures~\citep{stab2014identifying}, improving the persuasiveness of arguments~\citep{xia2022persua}, and teaching argumentation skills~\citep{wambsganss2021arguetutor}, 
while computational ideology analysis typically classifies perspectives~\citep{bamman2014,hardisty2010modeling} rather than generating perspective-aware reasoning.

We present TruthSplit, a comparative reasoning platform that reveals how different worldviews interpret the same arguments\footnote{The code is licensed under the MIT license and available at \url{https://github.com/unisg-ics-dsnlp/truthsplit}. A video demonstration is available at:\\\url{https://www.youtube.com/watch?v=xYMCOpU8x18}.}. TruthSplit combines a multi-stage analysis pipeline with a structured knowledge base encoding six ideological perspectives (Libertarian, Religious-Conservative, Ecological Social-Democrat, Populist-Nationalist, Communist, Neo-Reactionary) with their core values, conceptual definitions, and decision frameworks. The TruthSplit tool includes three-layer consistency testing based on Natural Language Inference (NLI), semantic concept linking, and an LLM-based worldview analysis. 
It supports both local, privacy-preserving models and larger proprietary LLMs available through external providers via API access to improve the quality of the analysis. To help users examine perspective-dependent disagreement, the system provides interactive visualizations that reveal worldview differences, reasoning trajectories, and patterns of divergence.
Figure \ref{fig:conceptual_overview} illustrates the central mechanism of TruthSplit: holding premises fixed while varying perspective priors to generate and analyze conditional reasoning. TruthSplit is intended for educators, students, researchers, and analysts seeking to explore perspective-dependent reasoning in political and ethical discourse.



Our contributions are fourfold:
(i) Operationalizing conditional validity: we enable users to perform a systematic analysis of arguments across multiple perspectives rather than assigning a single correctness label.
(ii) Perspective-conditioned reasoning: TruthSplit generates explicit reasoning chains conditioned on structured ideological profiles.
(iii) Interactive exploration of divergence: we provide a visual and analytical tool that identifies value conflicts, assumption gaps, and points of disagreement.
(iv) Structured worldview knowledge representation: we encode different perspectives as extensible computational profiles rather than informal prompt descriptions.


\section{Related Work}


TruthSplit builds on three strands of work: computational argumentation, interactive argumentation systems, and computational analyses of ideology and perspective. Existing approaches provide methods for extracting argumentative structures, assessing argument quality, supporting users in argument construction, or identifying ideological leaning in text. TruthSplit differs in that it uses structured worldview-specific background knowledge to condition the analysis of arguments. In that way, it allows users to compare how the same claim may be interpreted, evaluated, and explained differently across perspectives. 

\paragraph{Computational Argumentation}
NLP research has made significant advances in argument mining tasks \citep{lawrence-reed-2019-argument}, including the extraction of claims \citep{levy2014}, the identification of premises \citep{habernal2017}, and argument structure parsing \citep{stab2014identifying}. NLI has also been used to evaluate argument quality \citep{alkhatib2016}, and recent work has leveraged LLMs for generating arguments \citep{eskandari-miandoab-sarathy-2024-lets,huber-niklaus-2025-clear} and analyzing argument quality \citep{wachsmuth-etal-2024-argument,10.1007/978-3-031-63536-6_8}. These approaches provide important methods for identifying, structuring, and evaluating argumentative content. 
However, prior work generally does not incorporate structured background information about different perspectives into argument analysis, and therefore does not systematically explain how the same claim may yield different conclusions under different worldview-specific values, assumptions, and conceptual definitions.

\paragraph{Interactive Argumentation Systems}
Interactive systems support users in navigating, improving, or learning argumentation. CoArgue~\citep{liu2023coargue}, for example, extracts and summarizes argumentative elements in Community-Based Question Answering platforms to foster participation. Other systems visualize argument structures~\citep{huber-niklaus-2025-artist}, support persuasive writing~\citep{xia2022persua}, or teach argumentation skills through adaptive learning environments such as ArgueTutor~\citep{wambsganss2021arguetutor} and AL~\citep{wambsganss2020al}. TruthSplit similarly provides an interactive interface, but focuses on comparative analysis across ideological perspectives rather than participation support, persuasiveness, or argumentation training.

\paragraph{Worldview and Perspective Analysis}
Research in computational social science has explored ideological and perspective analysis, including methods for identifying political leanings in text~\citep{bamman2014}, analyzing ideological discourse \citep{20bb5633-6096-3ddf-b381-be41e519a74e}, and characterizing perspective differences \citep{hardisty2010modeling}. Such approaches are useful for detecting or categorizing perspectives, but typically do not generate perspective-conditioned reasoning about specific arguments. TruthSplit complements this work by representing worldviews as structured profiles of values, assumptions, definitions, and decision principles, and by using these profiles to support multi-perspective argument analysis.

\section{System Design \& Architecture}
\label{sec:design-architecture}

The objective of TruthSplit is to reveal ideological disagreement in arguments by operationalizing conditional validity, i.e., the idea that a conclusion may be valid given a specific set of normative priors, values, and conceptual definitions, even if it is rejected under alternative normative assumptions.
Existing argumentation systems typically assess structural quality \citep{wachsmuth-etal-2016-using} or classify ideological stance \citep{iyyer-etal-2014-political}. In contrast, TruthSplit addresses a different computational objective: analyzing how validity depends on perspective-specific priors. This is more similar to how humans interpret different texts, as every human holds some views, such as being more conservative or liberal leaning. For instance, under a strictly Libertarian worldview the claim that \textit{'All taxes are theft'} is rational, whereas a more socially-oriented individual would not support such a claim, and consider it to be too extreme to be considered rational. To achieve this, the system integrates structured perspective representations with logical inference and LLM-based reasoning in a unified comparative analysis framework.


Figure \ref{fig:architecture} provides an overview of the system architecture. TruthSplit consists of two main components: (i) a structured worldview knowledge base (see Section \ref{sec:worldview-kb}), and (ii) a multi-stage analysis pipeline that decomposes arguments, evaluates logical consistency, conditions reasoning on worldview profiles, and performs cross-perspective divergence analysis (see Section \ref{sec:pipeline}). The system is designed as an analytical and educational tool, i.e. it does not attempt to resolve disagreement, but to make explicit how different conclusions emerge from distinct normative starting points.




\subsection{Worldview Knowledge Base}
\label{subsec:worldviews}


\label{sec:worldview-kb}
We construct different worldviews based on established political philosophy literature and validate them through expert participants. We discuss the choice of worldviews further in Section \ref{sec:evaluation}. To cover a broad range of views and values, the worldviews are Libertarian, Religious-Conservative, Ecological Social-Democrat, Populist-Nationalist, Communist, and Neo-Reactionary, but these can be extended.\footnote{These profiles represent key positions across the political spectrum rather than an exhaustive catalog. The format is extensible, allowing further worldviews to be added manually.}
Each profile contains weighted core values, key definitions (i.e., how the worldview interprets contested terms), assumed principles, decision frameworks, and factor scores for 16 ideological dimensions. These factor scores make the worldviews computational, enabling quantitative comparison and systematic analysis across different perspectives. 
Since all profiles follow the same JSON schema, new or customized worldviews can be added without modifying the core analysis pipeline.
An example profile structure is included in Appendix~\ref{appendix:technical}. 

\subsection{Analysis Pipeline}
\label{sec:pipeline}

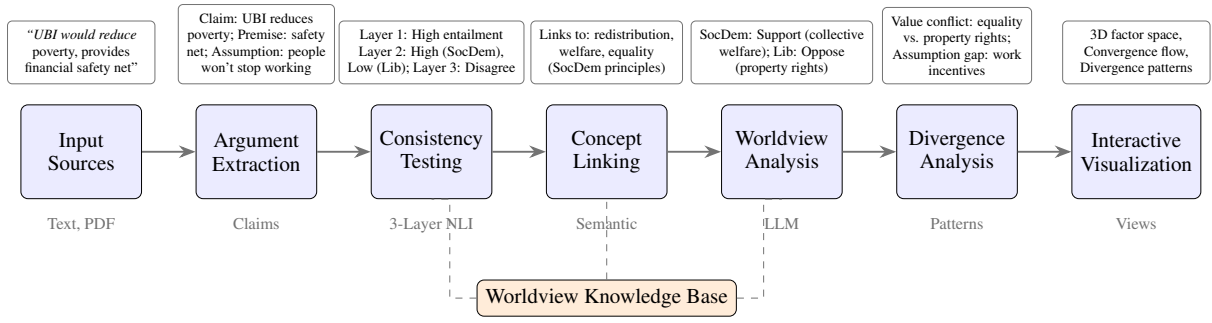
\begin{figure*}[t]
\centering
\resizebox{\linewidth}{!}{%
    \begin{tikzpicture}[
        node distance=0.8cm,
        stage/.style={rectangle, draw, rounded corners=3pt, minimum width=1.8cm, minimum height=1.4cm, align=center, fill=blue!8, font=\small},
        arrow/.style={-{Stealth[length=2.5mm]}, thick, color=gray!110},
        sublabel/.style={font=\scriptsize, text=gray!110, align=center},
        example/.style={rectangle, rounded corners=2pt, minimum width=2.2cm, minimum height=0.75cm, align=center, fill=white, font=\tiny, text=black, draw=gray!110, thin}
    ]
    \node[stage] (input) {Input\\Sources};
    \node[stage, right=of input] (nlp) {Argument\\Extraction};
    \node[stage, right=of nlp] (consistency) {Consistency\\Testing};
    \node[stage, right=of consistency] (linking) {Concept\\Linking};
    \node[stage, right=of linking] (analysis) {Worldview\\Analysis};
    \node[stage, right=of analysis] (divergence) {Divergence\\Analysis};
    \node[stage, right=of divergence] (viz) {Interactive\\Visualization};
    \draw[arrow] (input) -- (nlp);
    \draw[arrow] (nlp) -- (consistency);
    \draw[arrow] (consistency) -- (linking);
    \draw[arrow] (linking) -- (analysis);
    \draw[arrow] (analysis) -- (divergence);
    \draw[arrow] (divergence) -- (viz);
    \node[example, above=0.3cm of input] (ex-input) {\textit{``UBI would reduce}\\poverty, provides\\financial safety net''};
    \node[example, above=0.3cm of nlp] (ex-nlp) {Claim: UBI reduces\\poverty; Premise: safety\\net; Assumption: people\\won't stop working};
    \node[example, above=0.3cm of consistency] (ex-consistency) {Layer 1: High entailment\\Layer 2: High (SocDem),\\Low (Lib); Layer 3: Disagree};
    \node[example, above=0.3cm of linking] (ex-linking) {Links to: redistribution,\\welfare, equality\\(SocDem principles)};
    \node[example, above=0.3cm of analysis] (ex-analysis) {SocDem: Support (collective\\welfare); Lib: Oppose\\(property rights)};
    \node[example, above=0.3cm of divergence] (ex-divergence) {Value conflict: equality\\vs. property rights;\\Assumption gap: work\\incentives};
    \node[example, above=0.3cm of viz] (ex-viz) {3D factor space,\\Convergence flow,\\Divergence patterns};
    \node[sublabel, below=0.15cm of input] {Text, PDF};
    \node[sublabel, below=0.15cm of nlp] {Claims};
    \node[sublabel, below=0.15cm of consistency] {3-Layer NLI};
    \node[sublabel, below=0.15cm of linking] {Semantic};
    \node[sublabel, below=0.15cm of analysis] {LLM}; 
    \node[sublabel, below=0.15cm of divergence] {Patterns};
    \node[sublabel, below=0.15cm of viz] {Views};
    \node[rectangle, draw, rounded corners=3pt, minimum width=2.5cm, minimum height=0.5cm, align=center, fill=orange!15, font=\small, below=1.2cm of linking] (kb) {Worldview Knowledge Base};
    \draw[dashed, gray!110] (kb.north) -- (linking.south);
    \draw[dashed, gray!110] (kb.east) -- ++(0.4,0) |- (analysis.south);
    \draw[dashed, gray!110] (kb.west) -- ++(-0.4,0) |- (consistency.south);
    \end{tikzpicture}
}
\caption{Overview of the TruthSplit pipeline.}
\label{fig:architecture}
\end{figure*}

TruthSplit processes inputs through six sequential stages that transform raw arguments into comparative worldview analyses. Input can be provided via direct text entry, file upload, or dynamically fetched news articles from News API\footnote{\url{https://newsapi.org/}}. Figure~\ref{fig:architecture} shows an example claim being processed by all stages of the pipeline.

\subsubsection{Argument Extraction}

Natural language arguments are often implicit---claims embedded in narrative, premises unstated, assumptions hidden. TruthSplit decomposes text into \textit{claims} (central assertions), \textit{premises} (explicit evidence), and \textit{assumptions} (unstated connecting beliefs).
For an example, see box (2) in Figure \ref{fig:conceptual_overview}.
TruthSplit offers two extraction modes: (i) \textit{local extraction} using a sequence classification model ($\sim$75--80\% accuracy, fully privacy-preserving), and (ii) \textit{cloud-based extraction} via LLM with structured prompts and JSON schema validation ($\sim$95\%+ accuracy). See Appendix~\ref{appendix:technical} for model specifications and Appendix~\ref{appendix:prompts} for prompts.

\subsubsection{Consistency Testing}


To assess whether the argument is conditionally valid, we assess the logical coherence on three layers.\footnote{Note that these layers are a system design choice rather than a standard NLI taxonomy: Layer 1 isolates perspective-independent inferential support, Layer 2 conditions consistency on one worldview profile, and Layer 3 compares consistency scores across profiles to identify disagreement.} We use a model pre-trained on the MultiNLI dataset \cite{N18-1101}. See Appendix \ref{subsec:appendix-model-metadata} for model details. 

\paragraph{Layer 1 - Premise-Claim Logic} Do premises logically support the conclusion, independent of value judgments? Arguments failing here have fundamental logical flaws. A text could for instance contain multiple, disconnected arguments, or contain unsupported claims. An initial analysis aims to uncover badly structured arguments that do not provide value when analyzed further due to those fundamental flaws.

\paragraph{Layer 2 - Perspective-Internal} Does the claim align with a perspective's principles? A claim may be internally consistent within one framework but contradictory within another. The check makes use of the six worldview profiles (Section \ref{subsec:worldviews}) and analyzes the text through the respective perspective to confirm it is internally consistent with the worldview.

\paragraph{Layer 3 - Cross-Perspective} Do perspectives agree or disagree on this claim? High agreement suggests shared values; high disagreement indicates fundamental conflicts requiring analysis. The individual worldview profiles discussed in Section \ref{subsec:worldviews} are used to provide insight into where and why different perspectives agree or disagree.

\paragraph{}Continuing our UBI example (Figure~\ref{fig:architecture}): Layer 1 tests whether ``provides a financial safety net'' logically supports ``would reduce poverty'' (high entailment). Layer 2 reveals divergence: for Social Democrats, UBI aligns with their principle of collective welfare (high consistency), while for Libertarians, mandatory wealth redistribution conflicts with property rights (low consistency). Layer 3 confirms this fundamental disagreement across worldviews.


\subsubsection{Concept Linking and Worldview Analysis}
Certain concepts are viewed differently in certain worldviews (see box (4) in Figure \ref{fig:conceptual_overview}). For instance, the concept of \textit{freedom} means ``absence of coercion'' from a Libertarian perspective, but ``capability to flourish'' from Social Democrats' point of view. Each worldview contains concepts with a brief definition or description of what they mean in the context of the specific worldview. For a given input, we identify the most relevant concepts belonging to each worldview by calculating the cosine similarity between worldview-specific concept definition and the extracted claims. Model details and thresholds for the similarity calculation are included in Appendix \ref{appendix:technical}.

\begin{figure}[htbp]
\centering
\includegraphics[width=\columnwidth]{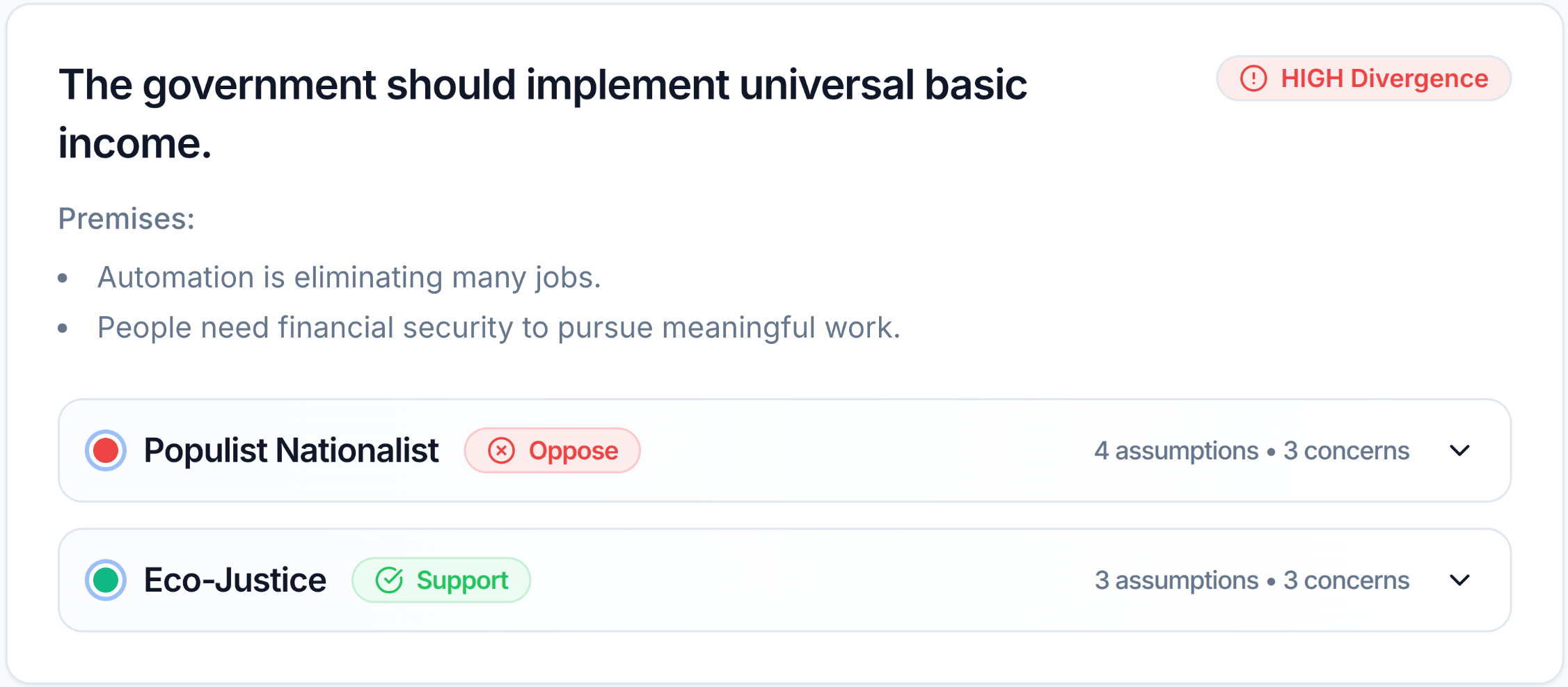}
\caption{Worldview reasoning showing positions for a UBI claim, with expandable assumptions and concerns. This figure shows only part of the complete reasoning output; the full analysis also includes logical chain, key assumptions, detailed concerns, alternative approach, and the metrics.}
\label{fig:reasoning-output}
\end{figure}

For worldview reasoning, we prompt LLMs with structured context including consistency scores, linked concepts, and full worldview profiles (Section~\ref{sec:worldview-kb}). Using structured prompts (Appendix~\ref{appendix:prompts}) and JSON schema validation, the LLM generates: interpretation, position (support/oppose/conditional), reasoning chain, key assumptions, concerns, and alternative approaches. Figure~\ref{fig:reasoning-output} shows a partial view of the output format.

\subsubsection{Divergence Analysis}


TruthSplit then provides an analysis into \textit{how} different perspectives disagree (see box (5) in Figure \ref{fig:conceptual_overview}). 
This is done on the levels of value conflicts (e.g., liberty vs. equality), which are caused by different prioritization issues and concepts, definition differences where the same concept is interpreted in different ways (e.g., ``rights'' as negative vs. positive claims), assumption gaps caused by a reliance on different empirical or normative assumptions, and priority differences where the overall values are shared or interpreted similarly but ranked differently (e.g., one might prioritize the need of the many over the good of the few, even though both agree on the overall underlying definitions). For the analysis we use LLMs with structured prompts to classify disagreement types and assess severity. Figure~\ref{fig:divergence} provides an example output.

The \textit{Convergence Flow} component complements this divergence analysis, and traces the reasoning chain from core values through beliefs, interpretation, and conclusion---showing at each step whether perspectives converge or diverge. This reveals where disagreement originates: do the different perspectives share values but differ in interpretation, or do they diverge from the start? The convergence flow is shown in Appendix Figure~\ref{fig:convergence-appendix}.

\section{User Interface}

The frontend is a multi-page React application with two main components: (i) an \textit{analysis dashboard} for claim extraction, worldview comparison, and result visualization, and (ii) an interactive \textit{worldview chatbot} for conversing with specific perspectives. 

\paragraph{Analysis Dashboard}
\label{sec:dashboard}

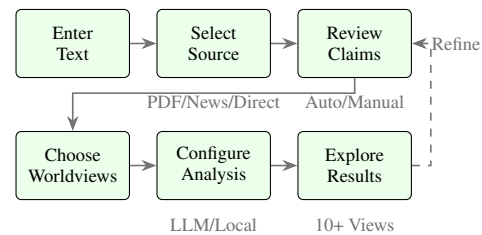
\begin{figure}[htbp]
\centering
\begin{tikzpicture}[
    node distance=0.5cm and 0.35cm,
    flowstep/.style={rectangle, draw, rounded corners=2pt, minimum width=1.5cm, minimum height=0.9cm, align=center, fill=green!8, font=\scriptsize},
    arrow/.style={-{Stealth[length=2mm]}, semithick, color=gray!110},
    choice/.style={font=\scriptsize, text=gray!110, align=center}
]
\node[flowstep] (input) {Enter\\Text};
\node[flowstep, right=of input] (source) {Select\\Source};
\node[flowstep, right=of source] (claims) {Review\\Claims};

\node[flowstep, below=0.7cm of input] (worldviews) {Choose\\Worldviews};
\node[flowstep, right=of worldviews] (config) {Configure\\Analysis};
\node[flowstep, right=of config] (results) {Explore\\Results};

\draw[arrow] (input) -- (source);
\draw[arrow] (source) -- (claims);
\draw[arrow] (claims.south) -- ++(0,-0.2) -| (worldviews.north);
\draw[arrow] (worldviews) -- (config);
\draw[arrow] (config) -- (results);

\node[choice, below=0.1cm of source] {PDF/News/Direct};
\node[choice, below=0.1cm of claims] {Auto/Manual};
\node[choice, below=0.1cm of config] {LLM/Local};
\node[choice, below=0.1cm of results] {10+ Views};

\draw[arrow, dashed, gray!110] (results.east) -- ++(0.25,0) |- node[pos=0.75, right, font=\scriptsize, text=gray!110] {Refine} (claims.east);
\end{tikzpicture}
\caption{User interaction flow from text input through claim extraction, worldview selection, and analysis configuration to interactive result exploration with 10+ visualizations.}
\label{fig:userflow}
\end{figure}

Figure~\ref{fig:userflow} illustrates the typical user interaction flow through the system. Users begin by providing input text through direct entry, PDF upload, or selecting from curated news articles. 
Next, the system extracts claims automatically. The user then chooses up to three worldviews for comparison of the claims and configures the analysis mode (external LLM provider, or local privacy-preserving model). 

\begin{figure}[H]
\centering
\includegraphics[width=\columnwidth]{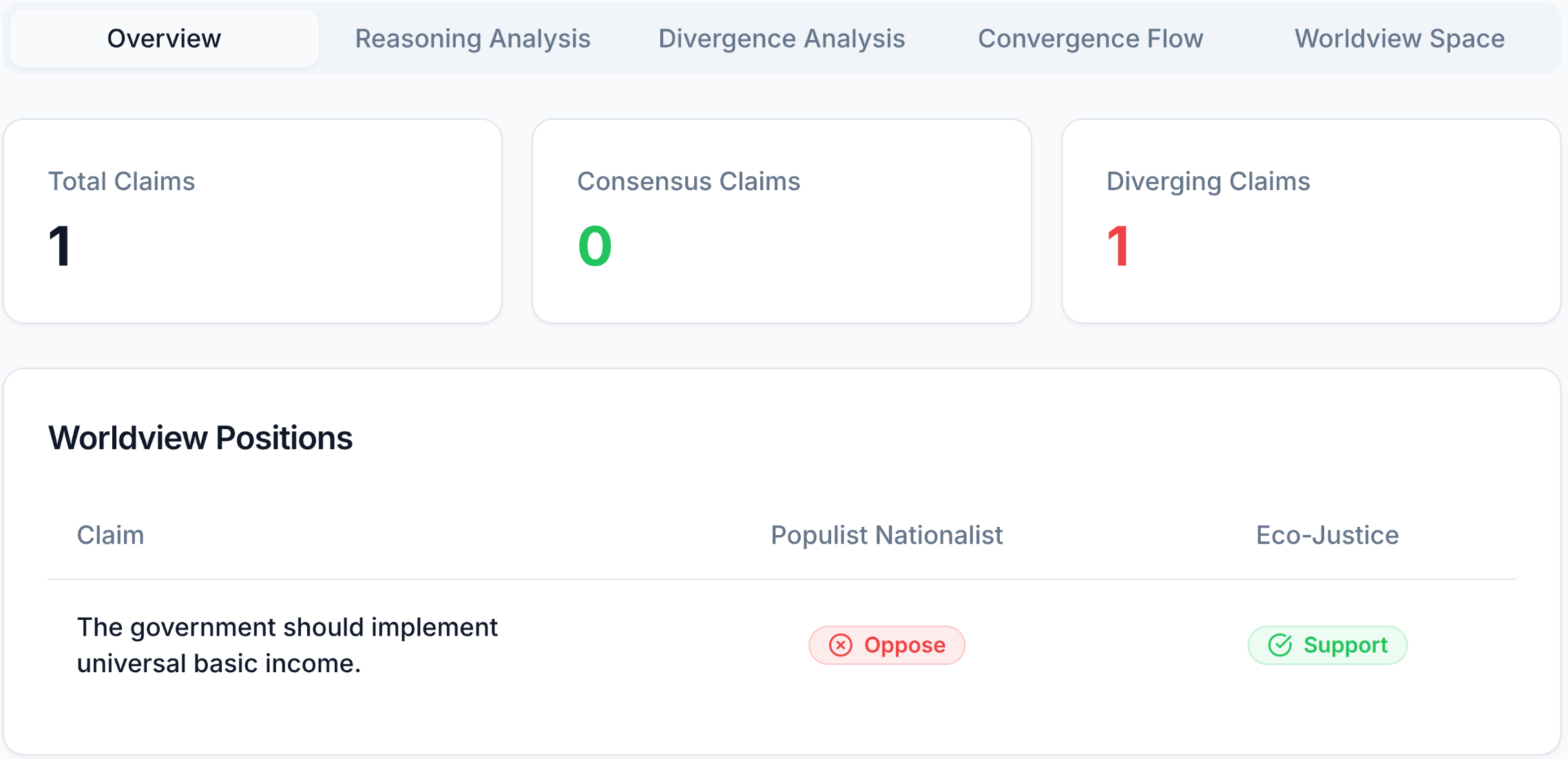}
\caption{Results dashboard: analysis summary, claims, worldviews, and position distribution.}
\label{fig:ui-results}
\end{figure}

The results interface provides multiple visualization tabs: Overview with summary metrics (Figure~\ref{fig:ui-results}), Detailed Analysis with per-worldview reasoning (Figure \ref{fig:reasoning-output}), Divergence Analysis showing disagreement patterns (Figure~\ref{fig:divergence}), Convergence Flow tracing reasoning chains (Appendix \ref{appendix:ui-screenshots} Figure~\ref{fig:convergence-appendix}), Worldview Space for factor positioning (Appendix \ref{appendix:ui-screenshots} Figure~\ref{fig:worldview-space-appendix}), and Worldview Chat for interactive dialogue.

\paragraph{Worldview Chatbot}
\label{sec:chatbot}


TruthSplit offers an interactive chatbot that embodies a selected worldview, enabling users to engage in dialogue and probe a perspective's reasoning in real-time. Users can ask follow-up questions, challenge positions, or explore how a worldview would respond to new scenarios---deepening understanding through conversational exploration.

The chatbot is based on an LLM, which can be configured through the config files, and assumes the position of a particular perspective profile, e.g. a conservative view. The user can then discuss the analysis results with the chatbot operating under a certain perspective, or have a regular conversation with it, again while the chatbot is assuming a certain perspective (i.e. a smalltalk setting).

The complete worldview is included in the dialogue context (core values, definitions, principles, red lines) and the model is prompted to respond consistently from that perspective. We include an evaluation of quality of the dialogues, as well as experiments to assess the stability, in Section~\ref{sec:evaluation}.

\section{Evaluation}
\label{sec:evaluation}



We conducted a mixed-methods evaluation to assess whether TruthSplit makes ideological disagreement more intelligible by exposing conditional validity across perspective frameworks. The evaluation focused on (i) usability and interpretability of comparative reasoning outputs, (ii) quality of structured worldview representations, and (iii) robustness across different LLMs. The focus of our evaluation is on the usefulness and interpretability of the analyses, not on the correctness of the overall reasoning or divergence explanations. We leave this for future work. 

\paragraph{Study Design} The study consisted of two tiers. An expert evaluation (n=3 philosophy students) assessed the coherence of each of the six worldviews, reasoning quality, and the validity of factor encodings across 16 ideological dimensions per worldview. A broader evaluation (n=52 participants from political science, computer science, psychology, business, and related academic backgrounds) focused on usability, accessibility, and clarity of the divergence analysis.
All participants were recruited from academic contexts and self-identified as politically centrist.

\paragraph{Usability and Accessibility}
Table \ref{tab:results_usability} summarizes usability results. Experts rated TruthSplit as easy to use (4.67/5) with high visual appeal (5.00/5). The divergence analysis was accessible to non-experts (4.07/5), which indicates that the comparative outputs can be understood without philosophical training. Option selection (i.e., local vs. cloud analysis modes) was less clearly understood (3.47/5), suggesting room for interface refinement. The quality of the claim extraction was rated moderately positively (6.67/10).

\begin{table}[!htb]
\centering
\small
\begin{tabular}{lcc}
\toprule
\textbf{Metric} & \textbf{Expert} & \textbf{Broader} \\
\midrule
Ease of use (1–5) & 4.67 & -- \\
Visual appeal (1–5) & 5.00 & 4.36 \\
Divergence understanding (1–5) & 4.33 & 4.07 \\
Options understanding (1–5) & -- & 3.47 \\
Claim extraction quality (1–10) & -- & 6.67 \\
\bottomrule
\end{tabular}
\caption{Usability and accessibility ratings.}
\label{tab:results_usability}
\end{table}

\paragraph{Worldview Representation Validation} Experts evaluated 96 factor–worldview combinations (16 dimensions across 6 worldviews). Implemented factor scores showed moderate positive correlation with expert importance ratings (r=0.33), with strongest alignment for Religious-Conservative and Ecological Social-Democrat worldviews (r=0.46).
Inter-expert variance was substantial with a mean standard deviation of 2.01 and 39\% of cases with disagreement $\geq$ 5 on a 1 to 10 scale. This suggests inherent ambiguity in quantifying ideological dimensions. 

Qualitative feedback consistently described Ecological Social-Democrat, Libertarian, and Communist representations as coherent and reflective of real-world political discourse. The divergence analysis was highlighted as particularly valuable for understanding why worldviews disagree.

\paragraph{LLM Robustness} The outputs that were generated using different families of LLMs (Claude, GPT, Gemini, Grok, DeepSeek) showed no significant quality differences according to both expert and broader participants. This suggests that TruthSplit's structured prompting approach standardizes worldview-conditioned reasoning across models.


\section{Discussion and Future Work}

The results suggest that TruthSplit succeeds in making ideological disagreement more interpretable by foregrounding conditional validity. Rather than resolving disagreement, the system reframes it as a product of differing premises and value hierarchies, supporting its positioning as an analytical and educational tool rather than a decision-making system.
The study also highlights design tensions: 2/3 expert participants struggled with worldview boundaries and definitions---a common challenge in computational approaches to ideology~\citep{grimmer2022text}. The exploratory nature and limited sample size mean findings should be interpreted as indicative rather than conclusive. Future evaluations with larger and more diverse participant groups will be necessary.

Several promising directions emerge: (i) a custom worldview builder allowing users to define their own ideological profiles, enabling analysis through perspectives not currently represented; 
(ii) educational integration with curriculum materials for teaching critical thinking and perspective-taking in civic education and journalism programs; and (iii) multimodal input support to extend analysis to audio and video sources.

\section{Conclusion}
TruthSplit addresses an important challenge in computational argument analysis: analyzing how the validity of an argument depends on worldview-specific normative priors. Rather than assigning a single correctness judgment, it reveals how conclusions emerge from distinct configurations of values, assumptions, and conceptual definitions.
By integrating argument mining, NLI-based consistency modeling, structured worldview representations, LLM-conditioned reasoning, and interactive visualization, TruthSplit operationalizes conditional validity as a computational objective. It enables systematic comparison of how different worldviews interpret the same argument while keeping the underlying factual premises fixed.
Our evaluation indicates that users can meaningfully interpret and trace these differences: the divergence analysis was accessible to non-experts, and expert validation demonstrated moderate alignment between implemented worldview encodings and domain intuition. These findings suggest that structured worldview conditioning provides a viable foundation for multi-perspective argument analysis.
TruthSplit illustrates how AI systems can move beyond binary correctness judgments towards a transparent analysis of normative disagreement. We hope this framework supports future research on perspective-aware NLP and fosters more informed engagement with ideological differences.

\section{Acknowledgements}
TruthSplit is part of the M-Rational project, which is funded by the Swiss National Science Foundation, under grant number 10004303. The SNF project URL is: \url{https://data.snf.ch/grants/grant/10004303}.
This system was developed as part of a student project at the University of St. Gallen. 

\bibliography{references}

\FloatBarrier
\appendix
\section{Technical Details}
\label{appendix:implementation}
\label{appendix:technical}

\subsection{Architecture and Stack}

TruthSplit follows a client-server architecture: a React/TypeScript frontend with Three.js (3D visualizations) and Plotly (charts), a Python Flask REST API backend, and containerized deployment via Docker. The system uses a unified API layer (OpenRouter) supporting multiple LLM providers (Claude, GPT, Gemini, Grok, DeepSeek).

\subsection{NLP Models}
\label{subsec:appendix-model-metadata}

\begin{itemize}
    \item \textbf{Local extraction and NLI}: facebook/bart-large-mnli (zero-shot classification, CPU-only, 2--5 sec/text, $\sim$75--80\% accuracy)
    \item \textbf{Cloud extraction}: OpenRouter API with structured JSON schema validation ($\sim$95\%+ accuracy)
    \item \textbf{Semantic embeddings}: all-MiniLM-L6-v2 (Sentence-BERT) for concept linking via cosine similarity
\end{itemize}

\subsection{Consistency Testing Thresholds}

Three-layer NLI testing categorizes cross-worldview agreement by variance in consistency scores: $\leq$0.05 (strong agreement), $\leq$0.15 (moderate agreement), $\leq$0.25 (moderate disagreement), $>$0.25 (strong disagreement).



\subsection{Worldview Profile Schema}

Each worldview is encoded as a JSON profile containing: \texttt{core\_values} (name, weight, description), \texttt{key\_definitions} (term, definition), \texttt{assumed\_principles} (name, priority, description), \texttt{warrants} (name, confidence, description), \texttt{red\_lines}, \texttt{evidence\_preferences}, \texttt{decision\_rule} (type, weight), and \texttt{extensions} including \texttt{factor\_scores} across 16 ideological dimensions.

\subsection{Performance}

Complete analysis time (typical: 1 claim, 2 worldviews): Local NLP + Cloud LLM: $\sim$12--25 seconds; Cloud NLP + Cloud LLM: $\sim$13--30 seconds. 

\section{Prompt Templates}
\label{appendix:prompts}

All prompts use structured system/user prompt pairs with JSON schema validation enforcing required fields, enumerated values, minimum content lengths, and nested object structures. Below is the worldview analysis prompt as a representative example; claim extraction, divergence analysis, convergence flow, and chatbot prompts follow similar patterns.

\textbf{System Prompt:} ``You are an expert in political philosophy and worldview analysis. Use the provided structured data to give precise, nuanced analysis of how different worldviews interpret claims. Respond with structured JSON.''

\textbf{User Prompt} (abbreviated): Given a claim, consistency scores (3 layers), worldview profile (core values, definitions), and linked concepts, the LLM generates: (1) interpretation (2--3 sentences), (2) position (support/oppose/conditional), (3) reasoning chain (3--5 logical steps from core values to conclusion), (4) detailed reasoning (3--5 sentences), (5) key assumptions (3--5 with related core values), (6) concerns (3--5 with explanations), and (7) alternative approach aligned with worldview values.


\section{Screenshots}

\begin{figure*}[!htbp]
    \centering
    \includegraphics[width=0.9\linewidth]{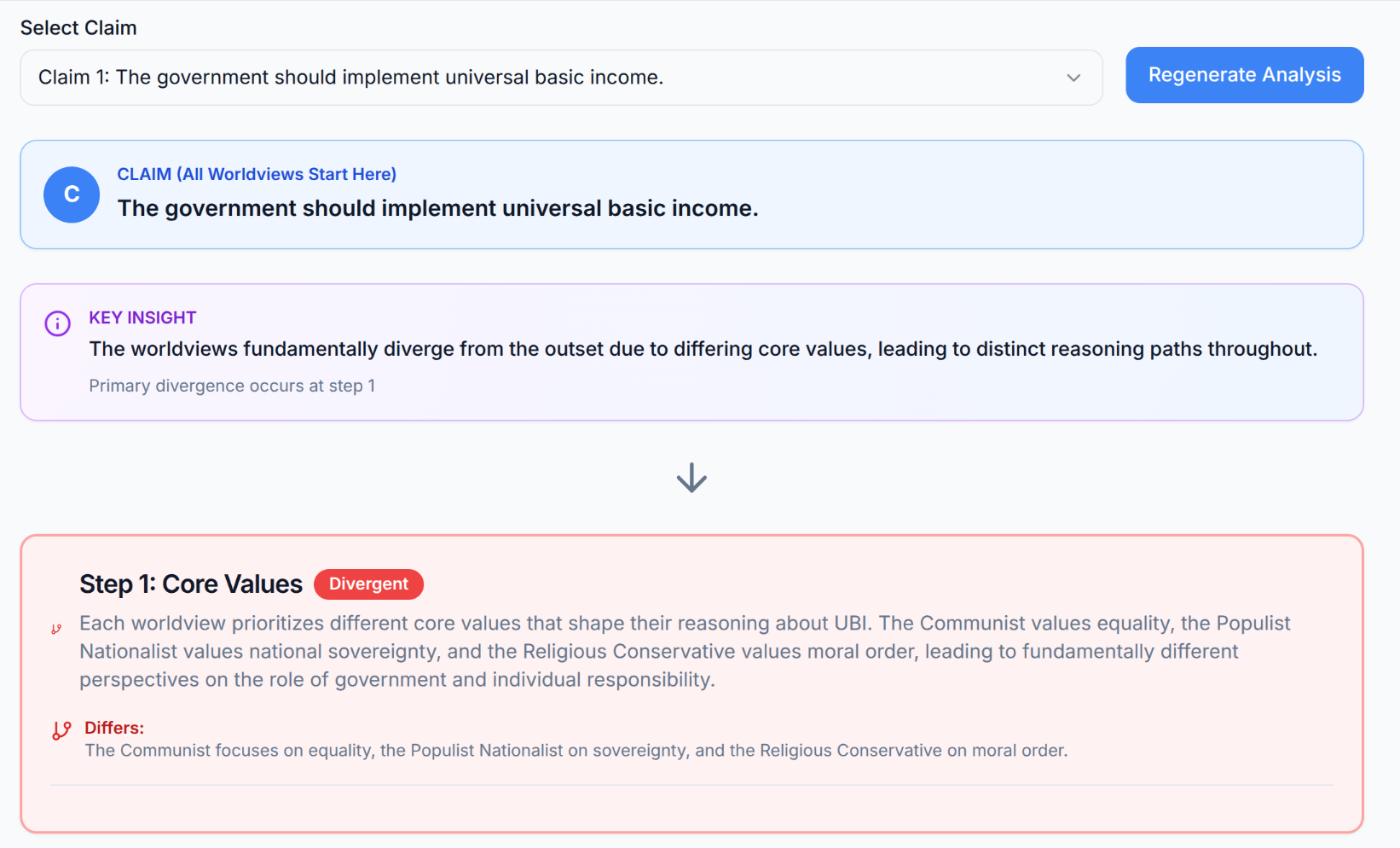}
    \caption{Excerpt for divergence flow analysis.}
    \label{fig:convergence-appendix}
\end{figure*}

\begin{figure*}[!htbp]
    \centering
    \includegraphics[width=0.9\linewidth]{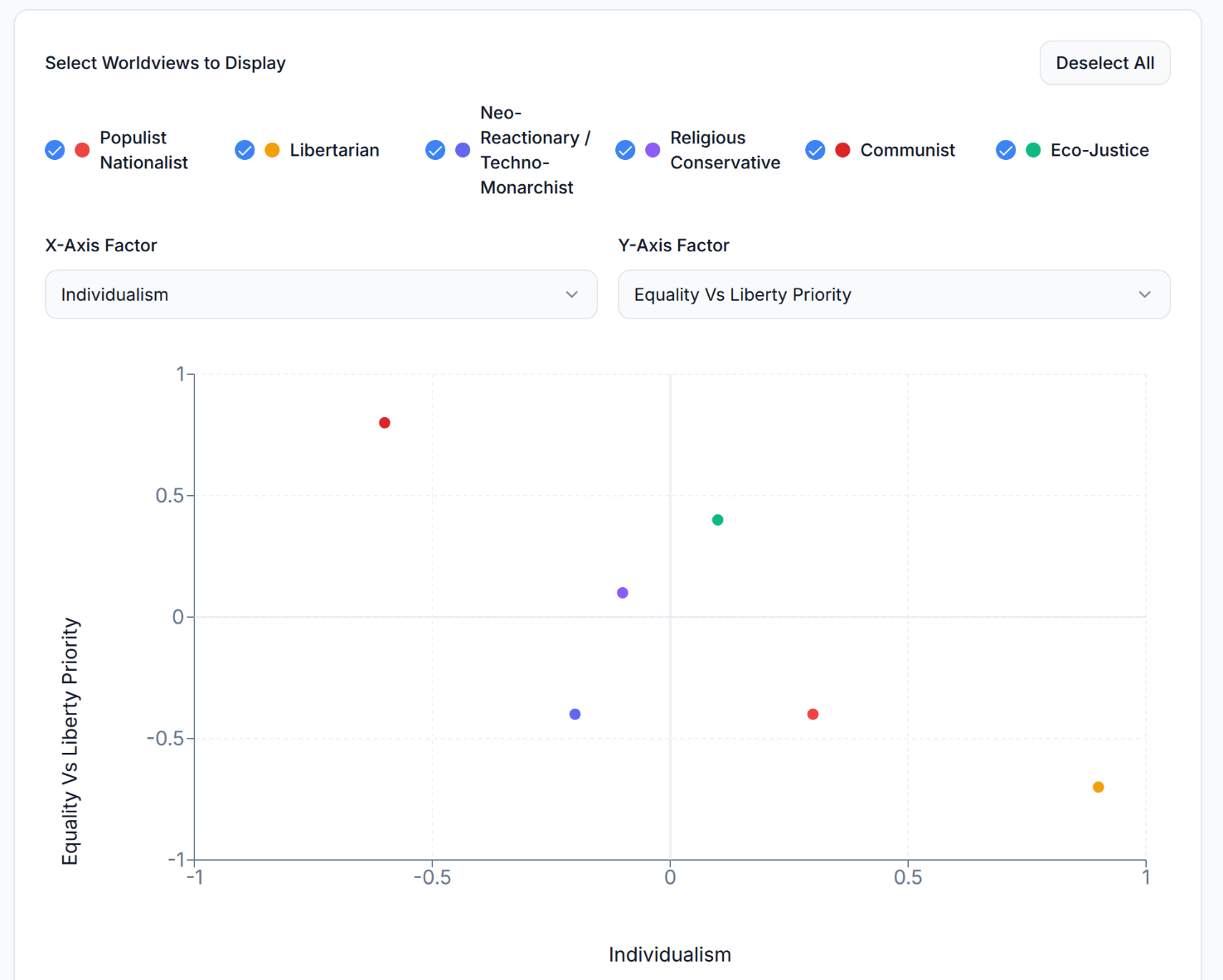}
    \caption{Screenshot of Worldview Positioning.}
    \label{fig:worldview-space-appendix}
\end{figure*}

\begin{figure*}[!htbp]
\centering
\includegraphics[width=0.8\linewidth]{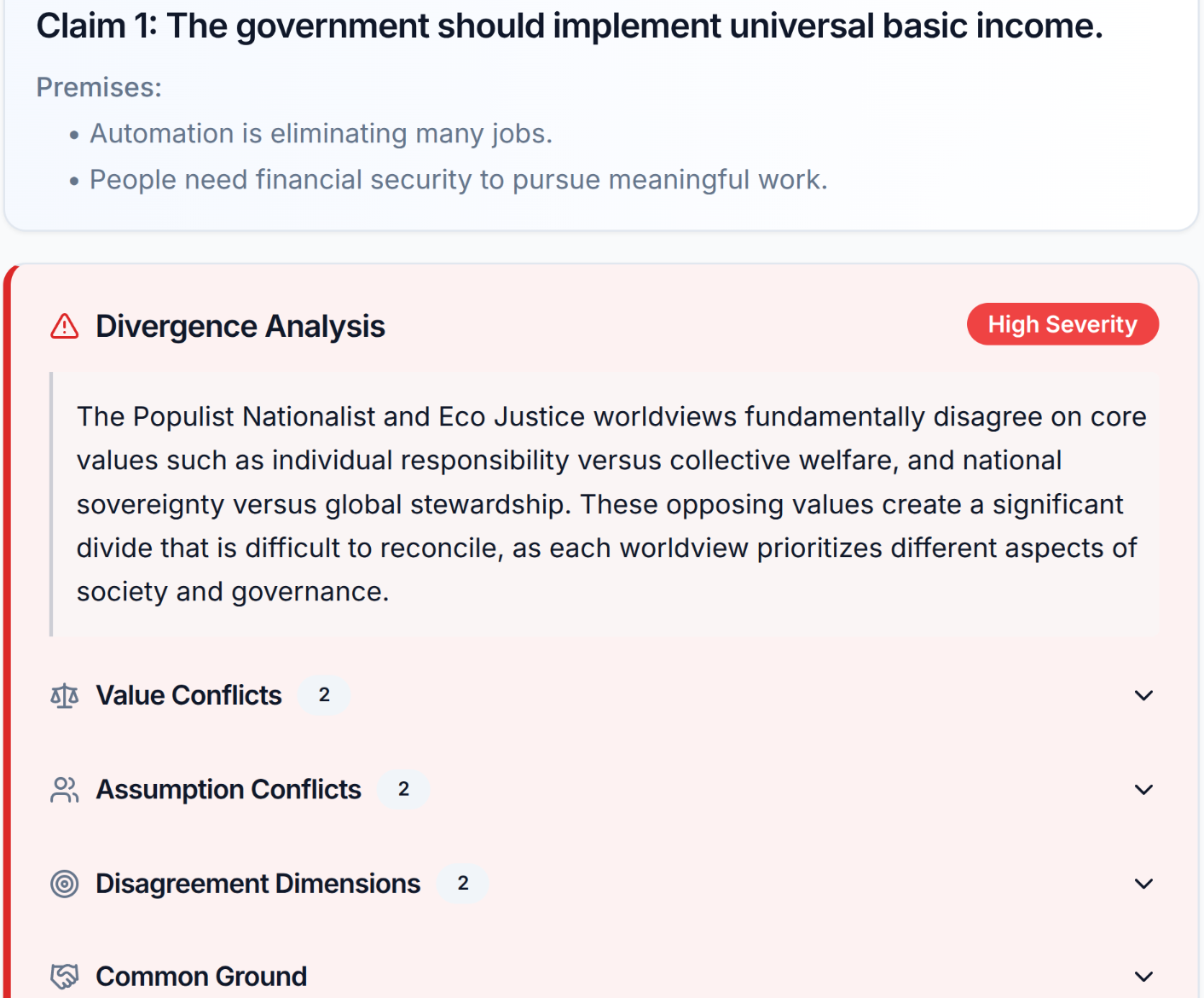}
\caption{Divergence Analysis showing value conflicts, assumption differences, and disagreement severity between worldviews.}
\label{fig:divergence}
\end{figure*}

Figures~\ref{fig:convergence-appendix}, \ref{fig:worldview-space-appendix}, and \ref{fig:divergence} illustrate the TruthSplit interface through representative screenshots.

\label{appendix:ui-screenshots}

\end{document}